\documentclass{article}

\usepackage{microtype}
\usepackage{graphicx}
\usepackage{subfigure}
\usepackage{booktabs} %

\usepackage{hyperref}

\usepackage[accepted]{icml2024}

\usepackage{amsmath}
\usepackage{amssymb}
\usepackage{mathtools}
\usepackage{amsthm}

\usepackage[capitalize,noabbrev]{cleveref}

\usepackage{tikz}
\usetikzlibrary{arrows.meta, positioning}

\usepackage{comment}

\usepackage{amsmath,amsfonts,bm}

\newcommand{\SNR}{\ensuremath{\operatorname{SNR}}}

\newcommand{\clip}{\ensuremath{\operatorname{clip}}}
\newcommand{\concat}{\ensuremath{\operatorname{concat}}}

\newcommand{\win}{\text{win}}
\newcommand{\clean}{\text{clean}}
\newcommand{\noise}{\text{noise}}
\newcommand{\lin}{\text{lin}}
\newcommand{\cln}{\text{cln}}
\newcommand{\init}{\text{init}}
\newcommand{\loc}{\text{loc}}

\def\eqref#1{equation~\ref{#1}}

\def\1{\bm{1}}

\def\eps{{\epsilon}}

\def\rmF{{\mathbf{F}}}

\def\rmI{{\mathbf{I}}}

\def\rmSigma{{\boldsymbol{\Sigma}}}

\def\veps{{\bm{\eps}}}

\def\vf{{\bm{f}}}

\def\vx{{\bm{x}}}

\def\vz{{\bm{z}}}

\DeclareMathAlphabet{\mathsfit}{\encodingdefault}{\sfdefault}{m}{sl}
\SetMathAlphabet{\mathsfit}{bold}{\encodingdefault}{\sfdefault}{bx}{n}

\def\sR{{\mathbb{R}}}

\newcommand{\E}{\mathbb{E}}

\newcommand{\R}{\mathbb{R}}

\newcommand{\KL}{D_{\mathrm{KL}}}

\theoremstyle{plain}

\theoremstyle{definition}

\theoremstyle{remark}

\usepackage[textsize=tiny]{todonotes}

\icmltitlerunning{Rolling Diffusion Models}

\begin{document}

\twocolumn[
\icmltitle{Rolling Diffusion Models}

\icmlsetsymbol{student}{*}

\begin{icmlauthorlist}
\icmlauthor{David Ruhe}{goog,uva,student}
\icmlauthor{Jonathan Heek}{goog}
\icmlauthor{Tim Salimans}{goog}
\icmlauthor{Emiel Hoogeboom}{goog}
\end{icmlauthorlist}

\icmlaffiliation{goog}{Google Deepmind, Amsterdam, Netherlands}
\icmlaffiliation{uva}{University of Amsterdam, Netherlands}

\icmlcorrespondingauthor{David Ruhe}{david.ruhe@gmail.com}
\icmlcorrespondingauthor{Jonathan Heek, Tim Salimans, Emiel Hoogeboom}{\{jheek, salimans, emielh\}@google.com}

\icmlkeywords{Machine Learning, ICML}

\vskip 0.3in
]

\printAffiliationsAndNotice{$^*$Work done as a Student Researcher at Google.}  %

\begin{abstract}
Diffusion models have recently been increasingly applied to temporal data such as video,
fluid mechanics simulations, or climate data.
These methods generally treat subsequent frames equally regarding the amount of noise in the diffusion process.
This paper explores \emph{Rolling Diffusion}: a new approach that uses a sliding window denoising process.
It ensures that the diffusion process progressively corrupts through time by assigning more noise to frames that appear later in a sequence, reflecting greater uncertainty about the future as the generation process unfolds.
Empirically, we show that when the temporal dynamics are complex, Rolling Diffusion is superior to standard diffusion.
In particular, this result is demonstrated in a video prediction task using the Kinetics-600 video dataset and in a chaotic fluid dynamics forecasting experiment.

\end{abstract}

\section{Introduction}
\label{sec:introduction}
Diffusion models \cite{sohldickstein2015diffusion,song2019generativemodellingestimatinggradient, ho2020denoising} have significantly boosted the field of generative modeling.
They provided the fundaments for large-scale text-to-image systems like DALL-E 2 \cite{ramesh2022hierarchicaltextconditional}, Imagen \cite{saharia2022imagen}, Parti \cite{yu2022scalingautoregressive}, and Stable Diffusion \cite{rombach2022highresolution}.
Other applications of diffusion models include density estimation, text-to-speech, and image editing \cite{kingma2021vdm, gao2023e3, kawar2023imagic}.

After these successes in these domains, interest in developing diffusion models for time sequences has grown.
Prominent recent large-scale works include, e.g., Imagen Video \cite{ho2022imagen}, Stable Diffusion Video \cite{StabilityAI2023StableVideo}.
Other impressive results for generating video data have been achieved by, e.g., \cite{blattmann2023align, ge2023preserve,harvey2022flexible,singer2022makeavideo,ho2022video}.
Applications of sequential generative modeling outside video include, e.g., fluid mechanics or weather and climate modeling \cite{price2023gencast,meng2022ondistillation,lippe2023pde}.

What is common across many of these works is that they treat the temporal axis as an `extra spatial dimension'.
That is, they treat the video as a 3D tensor of shape $K \times H \times W$.
This has several downsides.
First, the memory and computational requirements can quickly grow infeasible if one wants to generate long sequences.
Second, one is typically interested in being able to \emph{roll out} the sampling process for a variable number of time steps.
Therefore, an alternative angle is a fully autoregressive approach by conditioning on a sequence of input frames and simulating a single output frame, which is then concatenated to the input frames,
upon which the recursion can continue.
In this case, one has to traverse the entire denoising diffusion sampling chain for every single frame, which is computationally intensive.
Additionally, iteratively sampling single frames leads to quick autoregressive error accumulation.
A middle ground can be found by jointly generating blocks of frames.
However, in this \emph{block-autoregressive} case, a diffusion model would use the same number of denoising steps for every frame.
This is suboptimal since, given a sequence of conditioning frames, the generative uncertainty about the next few is much lower than the frames further into the future.
Finally, both methods sample frames only jointly with earlier frames, which is potentially a suboptimal parameterization.

\begin{figure*}
    \centering
    \includegraphics[width=0.95\linewidth]{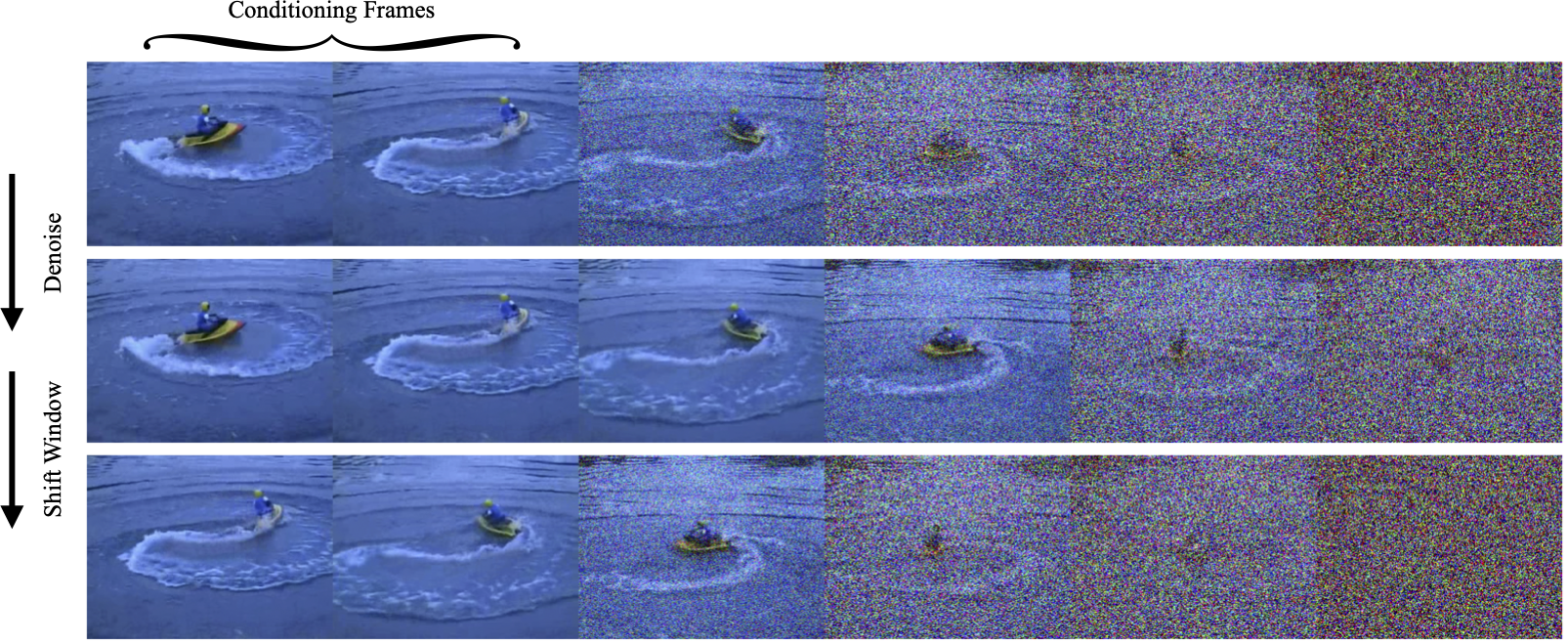}
    \caption{Overview of the Rolling Diffusion rollout sampling procedure. The input to the model contains some conditioning and a sequence of partially denoised frames. The model then denoises the frames by a small amount. After denoising, the sliding window shifts, and the fully denoised frames are concatenated with the conditioning. This process is repeats until the desired number of frames is generated. Example video taken from the Kinetics-600 dataset \cite{kay2017kinetics} (CC BY 4.0).}
    \label{supp:example_rolling}
\end{figure*}

In this paper, we propose a new framework called \emph{Rolling Diffusion}, a method that explicitly corrupts data from past to future.
This is achieved by reparameterizing the global diffusion time to a \emph{local time} for each frame.
It turns out that by doing this, one can (apart from boundary conditions) completely focus on a local sliding window sequential denoising process.
This has several temporal inductive biases, alleviating some of the abovementioned issues.
\begin{enumerate}
    \item In denoising diffusion models, the model output tends to contain low-frequency information in high noise regimes and includes high-frequency information only when corruptions are light. In our framework, the noise strength is higher for frames that are further from the conditioning. As such, the model only needs to predict low-frequency information (i.e., global structures) for frames further into the future; high-frequency information gets included as frames move closer to the present.
    \item Each frame is generated together with both a number of preceding and succeeding frames.
    \item Due to the local sliding window point of view, every frame enjoys the same inductive bias and undergoes a similar sampling procedure regardless of its absolute position in the video.
\end{enumerate}

These merits are empirically demonstrated in, among others, a video prediction experiment using the Kinetics-600 video dataset \cite{kay2017kinetics} and in an experiment involving chaotic fluid mechanics simulations.

\begin{figure*}
    \centering
    \includegraphics[width=\linewidth]{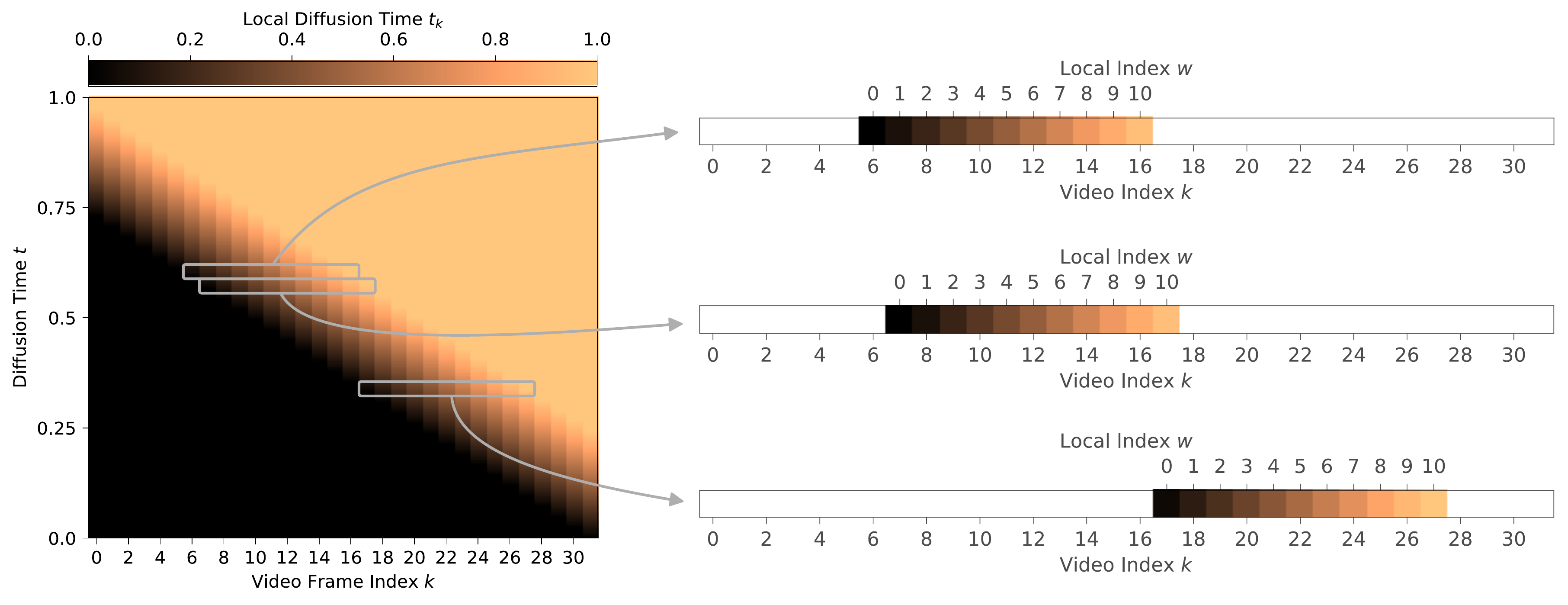}
    \caption{Left: an illustration of a global rolling diffusion process and its local time reparameterization.
        The global diffusion denoising time $t$ (vertical axis) is mapped to a local time $t_k$ for a frame $k$ (horizontal axis).
        The local time is then used to compute the diffusion parameters $\alpha_{t_k}$ and $\sigma_{t_k}$.
        On the right, we show how the same \emph{local} schedule can be applied to each sequence of frames based on the frame index $w$.
        The nontrivial part of sampling the generative process only occurs in the sliding window as it gets shifted over the sequence.
    }
    \label{fig:global_schedule}
\end{figure*}
\section{Background: Diffusion Models}
\label{sec:background}

\subsection{Diffusion}
Diffusion models consist of a process that destroys data stochastically,
named the `diffusion process', and a generative process called the denoising process.
Let $\vz_t \in \sR^D$ denote a latent variable over a diffusion dimension $t \in [0, 1]$.
We refer to $t$ as the \emph{global (diffusion) time}, which will determine the amount of noise added to the data.
Given a datapoint $\vx \in \sR^D$, $\vx \sim q(\vx)$, the diffusion process is designed so that $\vz_0 \approx \vx$ and $\vz_1 \sim \mathcal{N}(0, 1)$ via the distribution:
\begin{equation}
    q(\vz_t | \vx) := \mathcal{N}(\vz_t | \alpha_t \vx, \sigma_t^2 \rmI)\,,
    \label{eq:bg_diffusion}
\end{equation}
where $a_t$ and $\sigma_t^2$ are strictly positive scalar functions of $t$.
We define their \emph{signal-to-noise ratio}
\begin{align}
    \SNR(t) := \frac{\alpha_t^2}{\sigma_t^2} 
\end{align}
to be \emph{monotonically decreasing} in $t$.
Finally, we let $\alpha_t^2 + \sigma_t^2 = 1$, corresponding to a \emph{variance-preserving} process which also implies $a_t^2 \in (0, 1]$ and $\sigma_t^2 \in (0, 1]$.

Given the noising process, it can be shown \cite{sohldickstein2015diffusion} that the \textit{true} (i.e., optimal) denoising distribution for a single datapoint $\vx$ from time $t$ to time $s$ ($s \leq t)$ is given by
\begin{equation}
    q(\vz_s | \vz_t, \vx) = \mathcal{N}(\vz_s | \mu_{t \to s}(\vz_t, \vx), \sigma^2_{t \to s} \rmI) \,,
\end{equation}
where $\mu$ and $\sigma^2$ are \emph{analytical} mean and variance functions of $t$, $s$, $\vx$ and $\vz_t$.
The parameterized generative process $p_\theta(\vz_s | \vz_t)$ is then defined by approximating $\vx$ via a neural network $f_\theta: \sR^D \times [0, 1] \to \sR^D$.
That is, we set
\begin{equation}
    p_\theta(\vz_s | \vz_t) := q(\vz_s | \vz_t, \vx=f_\theta(\vz_t, t)) \,.
    \label{eq:bg_generative}
\end{equation}

The diffusion objective can be expressed as a KL-divergence between the diffusion process and the denoising process, i.e. 
$\KL(q(\vx, \vz_0, \ldots, \vz_1) \,||\, p(\vx, \vz_0, \ldots, \vz_1))$ which simplifies to \citep{kingma2021vdm}:
\begin{align}
    \mathcal{L}_\theta(\vx) & :=\mathbb{E}_{t \sim U(0, 1), \veps \sim \mathcal{N}(0, 1)} \left[ a(t) || \vx - f_\theta(\vz_{t, \veps}, t) ||^2\right] \notag \\
                            & \quad+ \mathcal{L}_{\mathrm{prior}} + \mathcal{L}_{\mathrm{data}}\,,
    \label{eq:bg_loss}
\end{align}
where $\mathcal{L}_\mathrm{prior}$ and $\mathcal{L}_{\mathrm{data}}$ are typically negligible.
The weighting $a(t)$ can be freely specified.
In practice, it was found that specific weightings of the loss result in better sample quality \citep{ho2020denoising}.
This is the case for, e.g., \emph{$\epsilon$-loss}, which corresponds to $a(t) = \mathrm{SNR}(t)$.

\subsection{Diffusion for temporal data}
If one is interested in generation of temporal data beyond typical hardware constraints, one must consider (autoregressive) conditional extension of previously generated data.
I.e., given an initial sample $\vx^{k}$ at a temporal index $k$, we want to sample a (faithful) conditional distribution $p(\vx^{k+1} | \vx^{k})$.
This process can then be extended to videos of arbitrary lengths.
As discussed in \Cref{sec:introduction}, it is not yet clear what kinds of parameterization choices are optimal to estimate this conditional distribution.
Further, no temporal inductive bias is typically baked into the denoising process.

\section{Rolling Diffusion Models}
\label{sec:rolling_diffusion}

We introduce \emph{rolling diffusion models}, merging the arrow of time with the (de)noising process.
To formalize this, we first have to discuss the global diffusion model.
We will see that the only nontrivial parts of the global process take place locally.
Defining the noise schedule locally is advantageous
since the resulting model does not depend on the number of frames $K$ and can be unrolled indefinitely.

\subsection{A global perspective}
Let $\vx \in \R^{D \times K}$ be a time series datapoint where $K$ denotes the number of frames and $D$ the dimensionality of each frame.
The core idea that allows rolling diffusion is a \emph{reparameterization} of the diffusion (denoising) time $t$ to a frame-dependent \emph{local (frame-dependent) time}: i.e.,
\begin{align}
    t \mapsto t_k \,.
\end{align}
Note that we still require $t_k \in [0, 1]$ for all ${k \in \{0, \dots, K-1\}}$.
Furthermore, we still have a monotonically decreasing signal-to-noise schedule, ensuring a well-defined diffusion process.
However, we now have a different signal-to-noise schedule for each frame.
In this work, we also always have
$t_k \leq t_{k+1}$,
i.e., the local denoising time of a given frame is smaller than the local time of the next frame.
This means we add more noise to future frames: a natural temporal inductive bias.
Note that this is not strictly required; one could also have a reverse-time inductive bias or a mixture.
An example of such a reparameterization is shown in \Cref{fig:global_schedule} (left).
We depict a map that takes a global diffusion time $t$ (vertical axis) and a frame index $k$ (horizontal axis), and computes a local time $t_k$, indicated with a color intensity.

\paragraph{Forward process}
We now redefine the forward process using the local time:
\begin{align}
    q(\vz_{t} | \vx) := \prod_{k=0}^{K-1} \mathcal{N}(\vz^k_t | \alpha_{t_k} \vx^k, \sigma_{t_k}^2 \rmI) \,,
\end{align}
where we can \emph{reuse} the $\alpha$ and $\sigma$ functions (now evaluated locally at $t_k$) from before.
Here, $\vx^k$ denotes the $k$-th frame of $\vx$.

\paragraph{True backward process and generative process}
Given a tuple $(s, t)$, $s \in [0, 1]$, $t \in [0, 1]$, $s \leq t$,
we can divide the frames $k \in \{0, \dots, K-1\}$ into three categories:
\begin{align}
    \clean(s, t) & := \{ k \mid s_k = t_k = 0 \}\,,                      \\
    \noise(s, t) & := \{ k \mid s_k = t_k = 1 \}\,,                      \\
    \win(s, t)   & := \{ k \mid s_k \in [0, 1), t_k \in (s_k, 1]  \} \,.
\end{align}
Note that here, $s$ and $t$ are both diffusion time-steps (corresponding to certain SNR levels), while $k$ denotes a frame index.
This categorization can be motivated using the schedule depicted in \Cref{fig:global_schedule}.
Given, for example, $t = 0.5$ and $s = 0.375$, we see that the first frame $k=0$ falls in the first category.
At this point in time, $\vz_{t_0} = \vz_{s_0}$ are identical given that $\lim_{t \to 0^+} \log \mathrm{SNR}(t) = \infty$.
On the other hand, the last frame $k=K-1$ (31 in the figure) falls in the second category, i.e.,
both $\vz_{t_{K-1}}$ and $\vz_{s_{K-1}}$ are distributed as independent standard Gaussians, given that $\lim_{t \to 1^-} \log \mathrm{SNR}(t) = -\infty$.
Finally, the frame $k=16$ falls in the third, most interesting category: the sliding window.
As such, observe that the \emph{true} denoising process can be factorized as:
\begin{align}
    q(\vz_{s} | \vz_{t}, \vx) = q(\vz_s^\clean | \vz_t, \vx) q(\vz_s^\noise | \vz_t, \vx) q(\vz_s^\win | \vz_t, \vx).
\end{align}
This is helpful because we will see that the only frames that need to be modeled are in the window.
Namely, the first factor has
\begin{align}
    q(\vz_{s}^{\clean} | \vz_{t}, \vx) = \prod_{k \in \clean(s, t)} \delta(\vz^k_s | \vz^k_t)\,.
\end{align}
In other words, if $\vz^k_t$ is already noiseless, then $\vz^k_s$ will also be noiseless. Regarding the second factor, we see that they are all independently normally distributed:
\begin{align}
    q(\vz_{s}^{\noise} | \vz_{t}, \vx) = \prod_{k \in \noise(s, t)} \mathcal{N} (\vz^k_s | 0, \rmI).
\end{align}
Simply put, in these cases $\vz^k_s$ is independent noise and does not depend on data at all.
Finally, the third factor has a true non-trivial denoising process:
\begin{align}
    q(\vz_{s}^{\win} | \vz_{t}, \vx) = \prod_{k \in \win(s, t)} \mathcal{N}(\vz^k_s | \mu_{t_k \to s_k}(\vz^k_t, \vx^k), \sigma_{t_k \to s_k}^2 \rmI) \notag
\end{align}
where $\mu_{t_k \to s_k}$ and $\sigma_{t_k \to s_k}^2$ are the analytical mean and variance functions.
Note that we can then optimally (w.r.t. a KL-divergence) factorize the generative process similarly:
\begin{align}
    p_\theta(\vz_{s} | \vz_{t}) := p(\vz_{s}^{\clean} | \vz_{t}) p(\vz_{s}^{\noise} | \vz_{t}) p_\theta(\vz_{s}^{\win} | \vz_{t})\,,
\end{align}
with $p(\vz_{s}^{\clean} | \vz_t) := \prod_{k \in \clean(s, t)} \delta(\vz^k_s | \vz^k_t)$ and $p(\vz_{s}^{\noise} | \vz_t) := \prod_{k \in \noise(s, t)} \mathcal{N} (\vz^k_s | 0, \rmI)$.
The only `interesting' parameterized part of the generative process then has
\begin{align}
    p_\theta(\vz_{s}^{\win} | \vz_t) & := \prod_{k \in \win(s, t)} q (\vz^k_s | \vz_t, \vx^k=f_\theta(\vz_t, t_k)).
\end{align}
In other words, we can \emph{only focus the generative process on the frames that are in the sliding window}.
Finally, note that we can choose to not condition the model on all $\vz^k_t$ that have $t_k$ = 0, since frames that are far in the past are likely to be independent of the current frame,
and this excessive conditioning would exceed computational constraints.
As such, we get
\begin{align}
    p_\theta(\vz_{s}^{\win} | \vz_t) & = 
                                       p_\theta(\vz_{s}^{\win} | \vz^{\clean}_t, \vz^{\win}_t) \\ &:\approx p_\theta(\vz_{s}^{\win} | \widehat{\vz_t^\clean}, \vz^{\win}_t)                 
\end{align}
where $\widehat{\vz_t^\clean}$ denotes a specific subset of $\vz^{\clean}_t$, typically including a few frames slightly before the current sliding window.

In \Cref{sec:rolling_diffusion_objective}, we motivate, in addition to the arguments above, the following loss function:
\begin{align}
    \mathcal{L}_{\win, \theta}(\vx)  := &\mathbb{E}_{t \sim U(0, 1), \veps \sim \mathcal{N}(0, 1)} \left[ L_{\win, \theta}(\vx; t, \veps) \right]
\end{align}
with
\begin{align}
    L_{\win, \theta} := \sum_{k \in \win(t)} a(t_k) || \vx^k - f^k_\theta(\vz_{t, \veps}^{\win}, \vz_{t, \veps}^{\clean}, t) ||^2\,, \notag
\end{align}
where we suppress some arguments for notational convenience.
Here, $\vz_{t, \veps}$ denotes a noised version of $\vx$ as a function of $t$ and $\veps$ and $a(t_k)$ is a weighting function leading to, e.g., the usual `simple' $\epsilon$-MSE loss, $v$-MSE loss, or $x$-MSE loss.

Observe \Cref{fig:global_schedule} again.
After training is completed, we can essentially sample from the generative model by traversing the image with the sliding window from the top left to the bottom right.

\subsection{A local perspective}
In the previous section, we discussed how rolling diffusion enables us to concentrate entirely on frames within a sliding window. Instead of using $t$
to represent the global diffusion time, which determines the noise level for all frames, we now redefine $t$
to determine the noise level for each frame in a smaller subsequence.
Specifically, running the denoising chain from $t=1$ to $t=0$ will sample a sequence such that the first frame is completely denoised, but the subsequent frames still retain some noise.
In contrast, the global process described earlier denoises an entire video.

Similar to before, we reparameterize $t$ to allow for different noise levels for each frame in the sliding window.
This reparameterization should be:
\begin{enumerate}
    \item \emph{local}, meaning we allow for sharing and reusing the parameterization across various positions of the sliding window, independent of their absolute locations.
    \item \emph{consistent under moving the window}, meaning that the noise level for the current frame $w$ when $t=0$ should match the noise level at $w-1$ at $t=1$. This consistency enables seamless denoising as the window slides, ensuring that each frame is progressively denoised while shifting positions.
    
\end{enumerate}
Let $W < K$ be the size of the sliding window, and $w \in \{0, \dots, W-1\}$ be the local indices of the frames.
To satisfy the first assumption, we define the schedule in terms of the local index $w$ (see \Cref{fig:global_schedule} (right)):
\begin{align}
    t \mapsto t^W_w\,.
\end{align}
For the second, we know $t^W_w$ must have
\begin{align}
    t^W_w = g\left( \frac{w + t}{W} \right)
    \label{eq:local_general}
\end{align}
for some monotonically increasing (in $t$) function $g: [0, 1] \to [0, 1]$.
We will sometimes suppress $W$ for notational convenience.
Note that due to the locality of the parameterization, the process can be unfolded indefinitely at test time.

\paragraph{A linear reparameterization}
In this work we typically put $g:= \mathrm{id}$, i.e.,
\begin{align}
    t_{w}^\lin = \frac{w + t}{W}\,.
\end{align}
See \Cref{fig:global_schedule} (right) for an illustration of how this local schedule is applied to each sequence of frames. Observe that $t_{w}^\lin \in [w / W, (w + 1) / W] \subseteq [0, 1]$. 
As such, we can directly use our SNR schedule to compute the diffusion parameters for each frame.

One can extend the linear local time to include clean conditioning frames.
Let $n_{\cln}$ denote the number of clean frames (chosen as a hyperparameter), then the local time for a frame $w$ is:
\begin{equation}
    t_{w}^\lin(n_{\cln}) := \clip \left( \frac{w + t - n_{\cln}}{W - n_{\cln}} \right)\,,
\end{equation}
where $\clip: \R \to [0, 1]$ clips value between 0 and 1.

\subsection{Boundary conditions}
While framing rolling diffusion purely from a local perspective is convenient for training and sampling, it introduces some complicated edge conditions.
Given the linear local time reparameterization $t^\lin_w$,
we have that given the diffusion time $t$ running from $1$ to $0$, the local times run from $(\frac{1}{W}, \frac{2}{W}, \ldots \frac{W}{W})$ to $(\frac{0}{W}, \frac{1}{W}, \ldots, \frac{W-1}{W})$.
This means that using this setting, the signal-to-noise ratios are never minimal for all frames, meaning we cannot start sampling from a completely noisy state.
Visually, in \Cref{fig:global_schedule}, the rolling sampling procedure can be seen as moving the sliding window over the diagonal from top left to bottom right such that the local times of the frames remain invariant upon shifting.
However, this means that placing the window at the very left edge still results in having partially denoised frames.

To account for this, we \emph{co-train} the rolling diffusion model with an additional schedule that can handle this boundary condition:
\begin{equation}
    t_{w}^\init := \clip \left( \frac{w}{W} + t \right) \,.
\end{equation}
This \textit{init} noise schedule can start from initial random noise and generates a video in the `rolling state'. 
That is, at diffusion time $t=1$, this will put all frames to maximum noise, and at $t=0$ the frames will be in the rolling state.
To be precise, it starts from local times $(1, 1, \ldots 1)$ and denoises to $(0, \frac{1}{W}, \frac{2}{W}, \dots, \frac{W-1}{W})$, after which we can start using the previously described $t_w^\lin$ schedule.
From a visual perspective, in \Cref{fig:global_schedule}, this corresponds to placing the window at the upper left corner and moving it down \emph{vertically}, until it reaches the stage where it can start moving diagonally.

On the interval $[0, \frac{1}{W}]$, this schedule contains the previous local schedule $t_{w}^\lin$ as a special case.
To see this, note that $t_w^\init(\frac 1W) = \left( \frac 1W, \frac 2W, \dots, 1 \right)$, which is the same as $t_w^\lin(1)$.
Similarly, one can check the case for $t_w^\init(0)$.
This means that the model could be trained solely with $t_{w}^\init$ to handle the boundaries as well as being able to roll out indefinitely.
At sampling time, one then still uses the $t_w^\init$ schedule but restricted to $[0, \frac{1}{W}]$.
The caveat, however, is that the $t^\lin_w$ schedule only gets selected $1 / W$ of the time during training (assuming $t \sim U(0, 1)$).
In contrast, this schedule is used almost exclusively at test time, with the exception being at the boundary.
As such, we find it beneficial to oversample the $t_{w}^\lin$ schedule during training based on a Bernoulli rate $\beta$.

\begin{figure*}
    \centering
    \includegraphics[width=\linewidth]{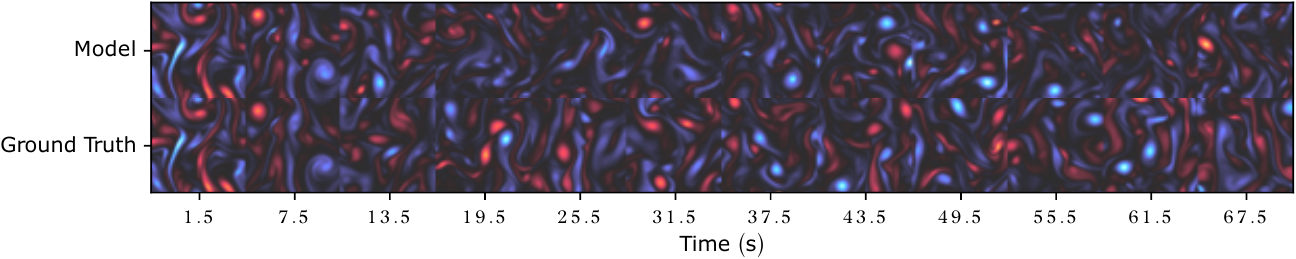}
    \caption{Sample Kolmogorov flow rollout. We observe that ground-truth structures are preserved initially, but the model diverges from the true data later on. Despite this, model is able to generate new turbulent dynamics much later on in the sequence.}
    \label{fig:kolmogorov_flow_rollout}
  \end{figure*}
\subsection{Local training}
We briefly discuss training details under the aforementioned local time reparameterizations.
We now consider $\vx \in \R^{D \times W}$, chunking videos into blocks of $W$ frames.
From $t_w$ one can compute $\alpha_{t_w}$ and $\sigma_{t_w}$ using typical SNR schedules.
Let $\vz \in \R^{D \times W}$, we have as the forward (noising) process
\begin{equation}
    \label{eq:local_diffusion_forward}
    q(\vz_{t} | \vx) := \prod_{w=0}^W \mathcal{N}(\vz^w_t | \alpha_{t_w} \vx^w, \sigma_{t_w}^2 \rmI),
\end{equation}
where $\vx^w$ denotes the $w$-th frame of $\vx$.

The training objective becomes
\begin{align}
    \mathcal{L}_{\loc, \theta}(\vx) := \mathbb{E}_{t \sim U(0, 1), \veps \sim \mathcal{N}(0, 1)} \left[ L_{\loc, \theta}(\vx; t, \veps) \right],
\end{align}
where we put
\begin{align}
    L_{\loc, \theta}(\vx, t, \veps) := \sum_{w=0}^W a(t_w) || \vx^w - f^w_\theta(\vz_{t, \veps}; t) ||^2\,.
\end{align}
\begin{algorithm}[t]
    \small
    \caption{Rolling Diffusion: Training}
    \label{alg:rolling_diffusion_training}
    \begin{algorithmic}
        \STATE {\bfseries Require:} $\mathcal{D}_\mathrm{tr} := \{\vx_1, \dots, \vx_N \}$, $\vx \in \R^{D \times W}$, $n_{\cln}$, $\beta$, $f_\theta$
        \REPEAT
        \STATE Sample $\vx$ from $\mathcal{D}_\mathrm{tr}$, $t \sim U(0, 1)$, $y \sim B(\beta)$
        \IF {$y$}
        \STATE Compute local time $t_{w}^\init(n_{\cln})$, $w=0, \dots, W-1$
        \ELSE
        \STATE Compute local time $t_{w}^\lin(n_{\cln})$, $w=0, \dots, W-1$
        \ENDIF
        \STATE Compute $\alpha_{t_w}$ and $\sigma_{t_w}$ for all $w=0, \dots, W-1$
        \STATE Sample $\vz_t \sim q(\vz_t | \vx)$ using \Cref{eq:local_diffusion_forward} (reparameterized from $\veps \sim \mathcal{N}(0, 1)$)
        \STATE Compute $\hat{\vx} \gets f_\theta(\vz_{t, \veps}; t)$
        \STATE Update $\theta$ using $L_{\loc, \theta}(\vx; t, \veps)$
        \UNTIL{Converged}
    \end{algorithmic}
  \end{algorithm}

\begin{algorithm}[t]
    \caption{Rolling Diffusion: Rollout}
    \small
    \label{alg:rolling_diffusion_unrolling}
    \begin{algorithmic}
        \STATE {\bfseries Require:} $p_\theta$, $n_{\cln}$, $\vz_0$ with local diffusion times $(0/W, \dots, (W-1)/W)$ (i.e., progressively noised).
        \STATE Video Prediction $\hat{\vx} \gets \{\vz_0^{n_{\cln}} \}$
        \REPEAT
        \STATE Sample $\vz^{W} \sim \mathcal{N}(0, \rmI)$
        \STATE $\vz_1 \gets \{\vz_0^1, \dots, \vz_0^{W-1}, \vz^{W}\}$
        \FOR{$t=1, (T-1)/T, \dots, 1/T$}
        \STATE Compute local times $t_w^\lin(n_{\cln})$,  $w =0, \dots, W-1$
        \STATE Sample $\vz_{t-1/T} \sim p_\theta(\vz_{t-1/T} | \vz_t)$
        \ENDFOR
        \STATE $\hat{\vx} \gets \hat{\vx} \cup \{\vz_0^{n_{\cln}}\}$
        \UNTIL{Completed}
    \end{algorithmic}
  \end{algorithm}
The training and sampling procedures are summarized in \Cref{alg:rolling_diffusion_training} and \Cref{alg:rolling_diffusion_unrolling}.
We summarize sampling at the boundary using $t^\init_w$ in \Cref{sec:algorithms}.
Furthermore, we provide a visual of the rolling sampling loop in \Cref{supp:example_rolling}.

\section{Related Work}
\label{sec:related_work}
\subsection{Video diffusion}
Video diffusion has been studied and applied directly in pixel space \cite{ho2022imagen,ho2022video,singer2022makeavideo} and in latent space \cite{blattmann2023align,ge2023preserve,he2022latent, yu2023video}, the latter typically empirically being slightly more effective.
Furthermore, these videos usually extend the two-dimensional image setting to three (two spatial dimensions and one temporal dimension) without considering autoregressive extension.

Methods that explore autoregressive video generation include \citet{yang2023diffusion,harvey2022flexible}. 
Directly parameterizing the conditional distribution of future frames given past frames is preferable \cite{harvey2022flexible,tashiro2021csdi} compared to adapting the denoising schedule of an unconditional diffusion model. 
Unlike previous approaches, Rolling Diffusion explicitly introduces a notion of time in the training procedure. 
\citet{harvey2022flexible} compare various conditioning schemes but do not explicitly consider a temporally adapted noise schedule.

\subsection{Other time-series diffusion models}
Apart from video, sequential diffusion models have also been applied to other modes of time-series data,
such as audio \cite{kong2021diffwave}, text \cite{li2022diffusion}, but also scientifically to weather data \cite{price2023gencast} or fluid mechanics \cite{kohl2023turbulent}.
\citet{lippe2023pde} show that incorporating a diffusion-inspired denoising procedure can help recover high frequency information that typically gets lost when using learned numerical PDE solver emulators.
Dyffusion \cite{cachay2023dyffusion} employs a forecasting and interpolation model in a two-stage fashion.
Both models are trained with MSE objectives. This means that the generative model can be interpreted as factorized Gaussians, potentially leading to blurry predictions when generating stochastic or highly chaotic data as considered in this work.
\citet{wu2023ar} also study autoregressive models with specialized noising schedules, focusing mostly on text generation.
Finally, \citet{zhang2023tedi} published around the same time as the current work a paper with similar ideas as rolling diffusion.
There are, however, some core differences.
This work analyzes in more detail how rolling diffusion relates to a global, well-defined diffusion process, motivating why training on a sliding window is acceptable.
We isolate the effect of rolling diffusion and study it in more detail, while \citet{zhang2023tedi} combine local noise levels with additional losses, potentially blurring the effect of the sliding window idea with the impact of these auxiliary terms.
Finally, we introduce rolling diffusion schedules that deal with the boundary situations, allowing for generating sequences in an end-to-end manner.

\section{Experiments}
\label{sec:experiments}
We conduct experiments using data from various domains and explore several conditioning settings.
In all our experiments, we use the \emph{Simple Diffusion} architecture \citep{hoogeboom2023simple} with equal parameters for both standard and rolling diffusion.
We use two-dimensional spatial convolution blocks after which we have transformer blocks that attend both spatially and temporally in the deepest layers.
Hyperparameter settings can be found in \Cref{sec:hyperparameters}.
A note on the runtime complexity can be found in \Cref{sec:runtime_complexity}.

\begin{figure}
  \centering
  \includegraphics[width=\linewidth]{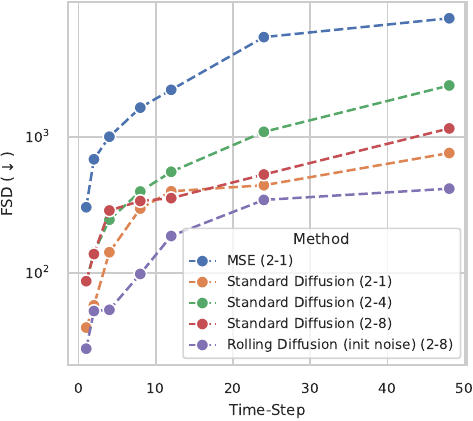}
  \caption{FSD results of the Kolmogorov Flow rollout experiment. Lower is better.}
  \label{fig:kolmogorov_flow}
\end{figure}
\begin{figure*}
  \centering
  \includegraphics[width=0.8\linewidth]{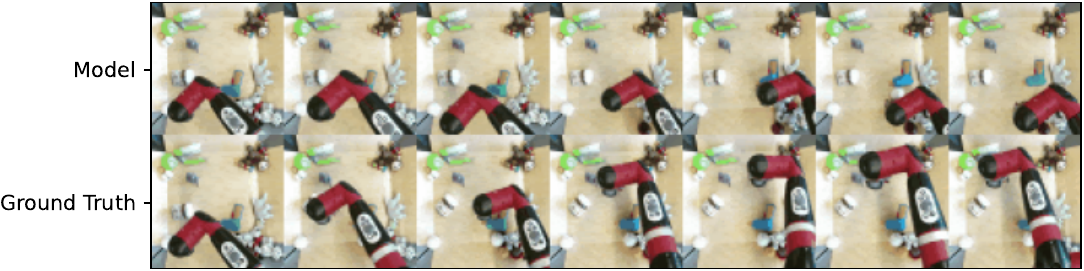}
  \caption{Top: Rolling Diffusion rollout on the BAIR Robot Pushing dataset. Bottom: ground-truth.}
  \label{fig:bair}
\end{figure*}

\subsection{Kolmogorov Flow}

First, we run an experiment on simulated fluid dynamics from JaxCFD \cite{Kochkov2021-ML-CFD,Dresdner2022-Spectral-ML}.
Specifically, we use the \emph{Kolmogorov flow}, an instance of the incompressible Navier-Stokes equations.
Recently, there has been increasing interest in emulating classical numerical PDE integrators with machine learning models.
Various results have shown that these have the capacity to simulate from initial conditions complex systems to high precision (e.g., \citet{li2020fourier}).
To similar ends, generative models are of increasing interest, as they provide several benefits.
First, they provide a way to directly obtain marginal distributions over a future state of a physical system, as opposed to numerically rolling out an ensemble of initial conditions.
This especially has use-cases in weather or climate modeling, fluid mechanics analyses, and stochastic differential equation studies.
Second, they can improve modeling high frequency information over approaches based on mean-squared error objectives.

The simulation is based on the following partial differential equation
$\frac{\partial \mathbf{u}}{\partial \tau} + \nabla \cdot (\mathbf{u} \otimes \mathbf{u}) = \nu \nabla^2 \mathbf{u} - \frac{1}{\rho} \nabla p + \mathbf{f}$, with
$\mathbf{u} : [0, \mathcal{T}] \times \mathbb{R}^2 \rightarrow \mathbb{R}^2$ is the solution, $\otimes$ the tensor product, $\nu$ the kinematic viscosity, $\rho$ the fluid density, $p$ the pressure field, and, finally, $\mathbf{f}$ the external forcing.
We randomly sample viscosities $\nu$ between $5 \cdot 10^{-4}$ and $5 \cdot 10^{-3}$, and densities between $0.5$ and $2$, making the task of predicting the dynamics non-deterministic.
The ground truth data is generated using a finite volume-based direct numerical simulation (DNS) with a maximal time step of $0.05$s with 6\,000 time-steps, corresponding to 300 seconds of simulation time, subsampled every 1.5s.
Note that this is much longer than typical simulations, which only consider simulation times in the order of tens of seconds.
We simulate 200\,000 such samples at $64 \times 64$, corresponding to 2 terabytes of data.
Due to the chaotic nature and the long simulation time, a model cannot be expected to predict the exact future state, which makes it an ideal dataset to test long temporal rollouts.
Details on the simulation settings can be found in \Cref{supp:simulation_details}.
The model is given 2 input frames containing the horizontal and vertical velocities.
Note that the long rollout lengths (up to $\pm$ 60 seconds), together with the large 1.5s strides, make this a more challenging task than the usual `neural emulator' task, where the predictions can be extremely accurate due to the shorter time-scales and higher temporal resolutions.
For example, \citet{lippe2023pde, sun2023neural} only go up to $\pm 15$s with much smaller strides.

\paragraph{Evaluation}
Typical PDE-emulation tasks directly compare models to ground truth using data-space RMSE scores.
However, in our uncertain setting we are more interested in how well the generated distribution matches the target distribution.
The ground-truth simulator provides no inherent uncertainty estimation, and running ensembles from various initial parameters is not straightforward due to a lack of proposal distribution for these initial settings.
Instead, we propose a method similar to \emph{Fr\'echet Inception Distance} (FID) or FVD.
We make use of the fact that spatial frequency intensities provide a good summary statistic \cite{sun2023neural,Dresdner2022-Spectral-ML,Kochkov2021-ML-CFD}.
Let $\vf_\vx \in \mathbb{R}^F$ denote a vector of spatial-spectral magnitudes computed using a (two-dimensional) \emph{Discrete Fourier Transform} (DFT) from a variable $\vx$.
Then, $\rmF_{\mathcal{D}} \in N \times F$ denotes all the frequencies of a (test-time) dataset $\mathcal{D}:= \{\vx_1, \dots, \vx_N \}$.
Let $\rmF_\theta$ denote the Fourier magnitudes of $N$ samples from a generative model.
We now compute the Fre\'chet distance between the generated samples and the true data by setting
\begin{align}
  &\mathrm{FSD}(\mathcal{D}, \theta) \notag \\ &:= \lVert \bar{\vf_{\mathcal{D}}} - \bar{\vf_\theta} \rVert^2  + \mathrm{tr}\Big{(} \rmSigma_{\mathcal{D}} + \rmSigma_\theta - 2 (\rmSigma_{\mathcal{D}} \rmSigma_{\theta})^{1/2} \Big{)}
\end{align}
where $\bar{\vf}$ and $\rmSigma$ denote the mean and covariance of the frequencies, respectively.
We call this metric the \emph{Fréchet Spectral Distance} (FSD).

We provide an example rollout in \Cref{fig:kolmogorov_flow_rollout}, where we plot the \emph{vorticity} of the velocity field.
Quantitatively, we present in \Cref{fig:kolmogorov_flow} the FSD results derived from the horizontal velocity fields of the fluid.
Note that the standard diffusion ($n_{\cln}$-1) baselines can be seen as a adaptations of PDE Refiner \citep{lippe2023pde} and \citep{kohl2023turbulent} for our task. 
Regarding rolling diffusion, we use the $t^\init_{w}(n_{\cln})$ reparameterization with $n_\cln=2$, and use $t^\lin_w(n_\cln)$ for long rollouts.
It is clear that an autoregressive MSE-based model, as typically used in the literature, is not suitable for this task.
For standard diffusion, we iteratively generate $W - n_{\cln}$ frames, after which we concatenate these to the conditioning and continue the rollout.
Rolling diffusion always shifts the window by one, sampling using the process described before.
Rolling diffusion consistently outperforms standard diffusion methods, regardless of conditioning settings and window sizes $(n_{\cln}, W-n_{\cln})$.
Additional qualitive and numerical results can be found in \Cref{supp:additional_results}.

\subsection{BAIR Robot Pushing Dataset}
\begin{table}[t]
  \small
  \centering
  \begin{tabular}{@{}lclll@{}}
    \toprule
    Method                                          & FVD ($\downarrow$) \\ \midrule
    DVD-GAN \cite{clark2019adversarial}             & 109.8              \\
    VideoGPT  \cite{yan2021videogpt}                & 103.3              \\
    TrIVD-GAN-FP                                    & 103.3              \\
    Transframer \cite{nash2022transframer}          & 100                \\
    CCVS     \cite{le2021ccvs}                      & 99                 \\
    VideoTransformer  \cite{weissenborn2019scaling} & 94                 \\
    FitVid    \cite{babaeizadeh2021fitvid}          & 93.6               \\
    NUWA  \cite{wu2022nuwa}                         & 86.9               \\
    Video Diffusion  \cite{ho2022video}             & 66.9               \\ \midrule
    Standard Diffusion (Ours)                       & \textbf{59.7}      \\
    Rolling Diffusion  (Ours)               & \textbf{59.6}      \\
    \bottomrule
  \end{tabular}
  \caption{Results of the BAIR Robot Pushing baseline experiment.}
  \label{tab:bair_1}
\end{table}

The Berkeley AI Research (BAIR) robot pushing dataset \cite{ebert2017self} is a standard benchmark for video {prediction}.
It contains 44\,000 videos at $64 \times 64$ of a robot arm pushing objects around.
Following previous methods, we condition in on 1 frame and predict the next 15.
We evaluate, consistently with previous works, using the \emph{Frech\'et Video Distance} (FVD) \cite{unterthiner2019fvd}.
For FVD, we use the I3D network \cite{carreira2017quo} by comparing $100 \times 256$ model samples against the 256 examples in the evaluation set.

Regarding rolling diffusion, we use the $t^\init_{w}(n_{\cln})$ reparameterization to sample the $W=16$ ($n_{\cln}=1$) frames to a partially denoised state, and then use $t^\lin_{w}(n_{\cln})$ to rollout and complete the sampling.
Note that standard diffusion samples all 15 frames at once and might be at an advantage since we do not consider autoregressive extension.

The results are shown in \Cref{tab:bair_1}.
We observe that both standard diffusion and rolling diffusion using the same (Simple Diffusion) architecture outperform previous methods.
Additionally, we see that there is no significant difference between the standard and rolling framework in this setting.
This is because the sampled sequences are, in both cases, indistinguishable from the true data \Cref{fig:bair}.

\subsection{Kinetics-600}
\begin{figure*}
  \centering
  \includegraphics[width=0.8\linewidth]{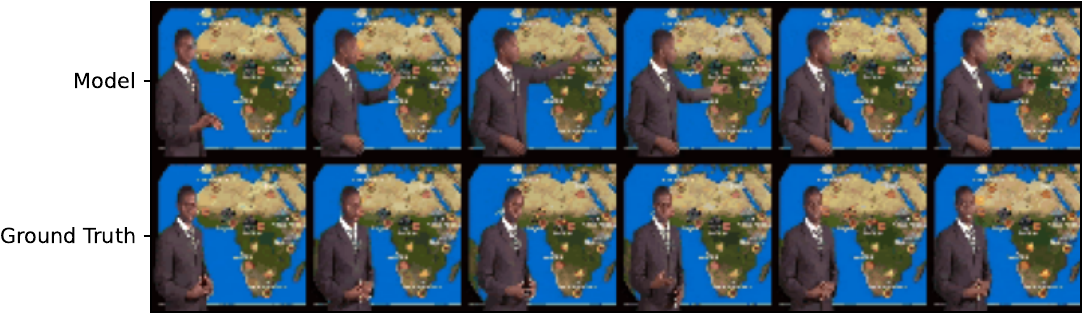}
  \caption{Top: Rolling Diffusion rollout on the Kinetics-600 dataset. Bottom: ground-truth. License: CC BY 4.0.}
  \label{fig:k600}
\end{figure*}

Finally, we evaluate video prediction on the Kinetics-600 benchmark \cite{kay2017kinetics,carreira2018short}.
It contains approximately 400\,000 training videos depicting 600 different activities rescaled to $64\times64$.
We run two experiments using this dataset.
The first is a baseline experiment in a setting equal to previously published works.
The next one specifically tests Rolling Diffusion's ability to autoregressively rollout for long sequences.

\paragraph{Baseline}
\begin{table}[t]
  \centering
  \small

  \begin{tabular}{@{}lc@{}}
    \toprule
    Method                                         & FVD ($\downarrow$)     \\ \midrule
    Phenaki  \cite{villegas2022phenaki}            & 36.4                   \\
    TrIVD-GAN-FP \cite{luc2020transformation}      & 25.7$^\dagger$         \\
    Video Diffusion \cite{ho2022video}             & 16.2                   \\
    RIN     \cite{jabri2022scalable}               & 10.8                   \\
    MAGVIT   \cite{yu2023magvit}                   & 9.9$^\dagger$          \\
    MAGVITv2    \cite{yu2023language}              & 4.3$^\dagger$          \\
    W.A.L.T.-L      \cite{gupta2023photorealistic} & \textbf{3.3}$^\dagger$ \\ \midrule
    Rolling Diffusion  (Ours)                            & $5.2$                  \\
    Standard Diffusion  (Ours)                       & \textbf{3.9}           \\
    \bottomrule
  \end{tabular}
  \label{tab:k600_1}
  \caption{FVD results of the Kinetics-600 baseline task (stride 1).
    Two-stage methods are indicated with `$\dagger$'.}
\end{table}
We compare against previous methods using 5 input frames and 11 output frames, and show the results in \Cref{tab:k600_1}
The evaluation metric is again FVD.
We note that many of the current SOTA methods are \emph{two-stage}, meaning that they use an autoencoder and run diffusion in latent space.
While empirically compelling \cite{rombach2022highresolution}, this makes it hard to isolate the effect of the diffusion model itself.
Note that it is not always clear whether the autoencoder parameters are included in the parameter count for two-stage methods, or on what data they are pretrained.
Running diffusion using the standard diffusion U-ViT architecture achieves an FVD of 3.9, which is comparable to the best two-stage methods.
Rolling diffusion has a strong disadvantage in this case: (1) the baseline generates all frames at once, thereby not suffering from autoregressive errors; and (2) with a stride of 1, there is very little dynamics in the 16 frames, mostly suitable for a standard diffusion model.
Still, rolling diffusion achieves a competitive FVD of 5.2.

\paragraph{Rollout}
\begin{table}[t]
  \small
  \centering
  \begin{tabular}{@{}llccc@{}}
    \toprule
                      & Method                         & Cond-Gen & FVD ($\downarrow$)                \\
    \midrule
    \textit{stride=8} & Standard Diffusion             & (5-11)   & 58.1               \\
    \textit{steps=24} & Rolling {\small $\beta=0.1$} & (5-11)   & \textbf{39.8 }     \\ \midrule
    \textit{stride=1} & Standard Diffusion             & (15-1)   & 1369\hspace{.15cm} \\
    \textit{steps=64} & Standard Diffusion             & (8-8)    & 157.1              \\
                      & Standard Diffusion             & (5-11)   & \textbf{123.7}     \\
                      & Rolling {\small $\beta=0.9$} & (5-11)   & 211.2              \\
    \bottomrule
  \end{tabular}
  \caption{Kinetics-600 (Rollout) with 8192 FVD samples, 100 sampling steps per 11 frames. Trained for 300k iterations. }
  \label{tab:k600_2}
\end{table}
Next, we compare the models' capabilities to autoregressively rollout and show the results in \Cref{tab:k600_2}.
All settings use a window size of $W=16$ frames, exploring several $n_{\cln}$ settings, denoted with `Cond-Gen'.
During training, we mix $t^\lin_w$ with a slightly adjusted version of $t^\lin_w$ (see \Cref{sec:rescaled_schedule}) at an oversampling rate of $\beta$.

We analyze two settings, one with a stride (also known as frame-skip or frame-step) of 1, rolling out for 64 steps, and another setting with a stride of 8 rolling out for 24 steps, effectively predicting ahead up to the $192$th frame.
In the first setting, standard diffusion performs better, quite possibly due to the invariability of the data.
We oversample the linear rolling schedule with a rate of $\beta=0.9$ to account for the high number of test-time steps.
Rolling diffusion consistently wins in the second setting, which is much more dynamic.
Note also that single-frame diffusion significantly underperforms here and that larger block autoregression is favorable.
See an example rollout in \Cref{fig:k600}.
Additionally, we compare to TECO \cite{yan2023teco}, which uses slightly different input-output settings, in \Cref{supp:additional_results}.
From \Cref{sec:beta_ablate}, we get an indication of the effect of the oversampling rate on the performance of rolling diffusion.
In this case, slightly oversampling $t^\lin_w$ yields the best result.

From these (and previous) results we draw the conclusion that rolling diffusion is particularly effective in dynamic settings, where the data is highly variable.

\section{Conclusion}
\label{sec:conclusion}
We presented \emph{Rolling Diffusion Models}, a new DDPM framework that progressively noises (and denoises) data through time.
Validating our method on video and fluid mechanics data, we observed that rolling diffusion's natural inductive bias gets most effectively exploited when the data is highly dynamic.
In this setting, Rolling Diffusion outperforms various existing methods.
This allows for exciting future directions in, e.g., video, audio, and weather or climate modeling.

\section*{Impact Statement}
Sequential generative models, including diffusion models, have a significant societal impact with applications in video generation and scientific research by enabling fast, highly detailed sampling. While they offer the upside of creating more accurate and compelling synthesis in fields ranging from climate modeling to medical imaging, there are notable downsides regarding content authenticity and originality of digital media.

\bibliography{iclr2022_conference.bib}
\bibliographystyle{icml2024}

\clearpage
\appendix
\onecolumn

\section{Rolling Diffusion Objective}
\label{sec:rolling_diffusion_objective}
\paragraph{Standard Diffusion}
For completeness, we briefly review the diffusion objective.
Let $T \in \mathbb{N}$ be a finite number of diffusion steps, and $i \in \{0, \ldots, T\}$ be a diffusion time-step.
\citet{kingma2021vdm} show that the discrete-time $\KL$ between $q(\vx, \vz_{0}, \dots, \vz_{T} )$ and $p(\vx, \vz_{0}, \ldots, \vz_{T})$ can be decomposed as
\begin{align}
  \KL \left( q(\vx, \vz_{0}, \dots, \vz_{T}) || p(\vx, \vz_{0}, \ldots, \vz_{T}) \right) &= \E_{q(\vx, \vz_{0}, \dots, \vz_{T})} \left[ \log q(\vx, \vz_{0}, \dots, \vz_{T}) - \log p(\vx, \vz_{0}, \ldots, \vz_{T}) \right] \\
  &=  c + \underbrace{\KL(q(\vz_T |\vx) || p(\vz_T))}_{\text{Prior Loss}} + \underbrace{\E_{q(\vz_0|\vx)} \left[ - \log p(\vx | \vz_0) \right]}_{\text{Reconstruction Loss}} + \mathcal{L}_D,
\end{align}
where $c$ is a data entropy term.
The prior and reconstruction loss terms are typically negligible.
$\mathcal{L}_D$ is the \emph{diffusion loss}, which is defined as
\begin{equation}
  \label{eq:discrete_diffusion_loss}
  \mathcal{L}_D = \sum_{i=1}^T \E_{q(\vz_{t_i} | \vx)} \left[ \KL ( q(\vz_{s_i} | \vz_{t_i}, \vx) || p(\vz_{s_i} | \vz_{t_i}) )\right]\,,
\end{equation}
Further, when $T \rightarrow \infty$ we get \emph{continuous} analog of \Cref{eq:discrete_diffusion_loss} \citep{kingma2021vdm}.
In practice, instead of using the resulting KL objective, one typically uses a weighted loss, which in some cases still corresponds to an importance-weighted KL \cite{kingma2023understanding}.
\begin{align}
  \mathcal{L}_w &:= \frac 12 \E_{t \sim U(0, 1), \veps \sim \mathcal{N}(0, 1)} \left[ w(\lambda_t) \cdot - \frac{d \lambda_t}{dt} \lVert \hat{\veps}_\theta(\vz_t; \lambda_t) - \veps \rVert^2 \right]\,,  \label{eq:continuous_diffusion_loss}
\end{align}
where $\lambda_t := \log \SNR(t)$. 
Note that in the main paper we directly write the weighting factor expression as $a(t)$ for ease of notation.
The weighting function is often changed to improve image quality, for instance by being $-1/\frac{d \lambda_t}{dt}$ so that the objective becomes the simple $\epsilon$-loss.

\paragraph{Rolling Diffusion Objective}
In rolling diffusion, the signal-to-noise schedule is unchanged but one has to account for the local time reparameterization. 
Recall that $t_k := t_k(t)$ denotes the local time reparameterization.
Using similar derivations as \cite{kingma2021vdm}, where we can factorize the KL divergence also over frame indices, we get the rolling diffusion loss:
\begin{align}
 \mathcal{L}_\infty &= \E_{t \sim U(0, 1), \veps \sim \mathcal{N}(0, 1)} \left[ \sum_{k=1}^K w(\lambda(t_k)) \cdot -\frac{d\lambda(t_k)}{d t_k} \frac{d t_k}{dt} \cdot \lVert \veps^k - \hat{\veps}^k_\theta(\vz_{\veps, t}; t) \rVert^2 \right]\,,
\end{align}
where we again can apply a custom weighting factor.

Recall the frame categorization of the main paper, i.e.,
\begin{align}
  \clean(s, t) & := \{ k \mid s_k = t_k = 0 \}\,,                      \\
  \noise(s, t) & := \{ k \mid s_k = t_k = 1 \}\,,                      \\
  \win(s, t)   & := \{ k \mid s_k \in [0, 1), t_k \in (s_k, 1]  \} \,,
\end{align}
we can see that $t'_k(t) = 0$ for $k \in \clean(t-dt, t)$ and $k \in \noise(t-dt, t)$ and thus the objective only has non-zero loss over the window:
\begin{align}
  \mathcal{L}_\infty &= \E_{t \sim U(0, 1), \veps \sim \mathcal{N}(0, 1)} \left[ \sum_{k \in \win(t)} w(\lambda(t_k)) \cdot -\frac{d\lambda(t_k)}{d t_k} \frac{d t_k}{dt} \cdot \lVert \veps^k - \hat{\veps}^k_\theta(\vz_{\veps, t}; t) \rVert^2 \right]\,.
\end{align}

Since the weighting function $w$ is arbitrary, we can choose it such that the entire prefactor vanishes, resulting in the typical $\epsilon$-loss.
However, the variational bound tells us that we should pay no price for any error made on the noise or clean frames.
Again, in the main paper we replace all the prefactors with $a(t_k)$ for notational convenience.

\section{Algorithms}
\label{sec:algorithms}
\begin{algorithm}[H]
  \label{supp:algorithms}
  \caption{Rolling Diffusion: sampling at the boundary.}
  \small
  \label{alg:rolling_diffusion_sampling}
  \begin{algorithmic}
      \STATE {\bfseries Require:} $\vx \in \R^{D \times n_{\cln}}$, $W$, $T$, $t_w^\init$, $p_\theta$ 
      \STATE Sample $\vz_1 \sim \mathcal{N}(0, \rmI)$, $\vz_1 \in \R^{D \times (W - n_{\cln})}$
      \STATE $\vz_1 \gets \concat(\vx, \vz_1)$
      \FOR{$t=1, (T-1)/T, \dots, 1/T$}
      \STATE Compute local times $t_{w}^\init(n_{\cln})$, $w=0, \dots, W-1$
      \STATE Sample $\vz_{t-1/T} \sim p_\theta(\vz_{t-1/T} | \vz_t)$
      \ENDFOR
      \STATE Return $\vz_0$, which now has local times $(0 / W, 1 / W, \dots, (W-1)/W)$
  \end{algorithmic}
\end{algorithm}

\section{Hyperparameters}
\label{sec:hyperparameters}
In this section we denote the hyperparameters for the different experiments. Throughout the experiments we use U-ViTs which are essentially U-Nets with MLP Blocks instead of convolutional layers when self-attention is used in a block. In the PDE experiments relatively small architectures are used. 
For BAIR, we used a larger architecture, increasing both the channel count and the number of blocks. For K600 we used even larger architectures, because this dataset turned out to be the most difficult to fit.

\begin{table}[H]
\centering
\scalebox{.9}{\begin{tabular}{@{}ll@{}}
\toprule
Parameter &  Value  \\ \midrule
Blocks & [3 + 3, 3 + 3, 3 + 3, 8] \\
Channels & [128, 256, 512, 1024] \\
Block Type & [Conv2D, Conv2D, Transformer (axial), Transformer] \\
Head Dim & 128 \\
Dropout & [0, 0.1, 0.1, 0.1] \\
Downsample & (1, 2, 2) \\
Model parametrization & $v$ \\
Loss & $\eps$-loss ($x$-loss with SNR weighting) \\
Number of Steps & 100\,000 (rollout experiments) / 570\,000 (standard experiments) \\
EMA decay & 0.9999 \\
learning rate & 1e-4 \\
\bottomrule
\end{tabular}}
\caption{Hyperparameters used for the Kolmogorov flow experiment.}
\label{tab:hyperparams_pde}
\end{table}

\begin{table}[H]
\centering
\scalebox{.9}{\begin{tabular}{@{}ll@{}}
\toprule
Parameter &  Value  \\ \midrule
Blocks & [4 + 4, 4 + 4, 4 + 4, 8] \\
Channels & [256, 512, 1024, 2048] \\
Block Type & [Conv2D, Conv2D, Transformer (axial), Transformer] \\
Head Dim & 128 \\
Dropout & 0.1 \\
Downsample & (1, 2, 2) \\
Model parametrization & $v$ \\
Loss & $\eps$-loss ($x$-loss with SNR weighting) \\
Number of Steps & 660\,000 \\
EMA decay & 0.9999 \\
learning rate & 1e-4 \\
\bottomrule
\end{tabular}}
\caption{Hyperparameters used for the BAIR robot pushing experiment.}
\label{tab:hyperparams_bair}
\end{table}

\begin{table}[H]
\centering
\begin{tabular}{@{}ll@{}}
\toprule
Parameter &  Value  \\ \midrule
Blocks & [4 + 4, 4 + 4, 5 + 5, 8] \\
Channels & [256, 512, 2048, 4096] \\
Block Type & [Conv2D, Conv2D, Transformer (axial), Transformer] \\
Head Dim & 128 \\
Dropout & [0, 0, 0.1, 0.1] \\
Downsample & (1, 2, 2) \\
Model parametrization & $v$ \\
Loss & $\eps$-loss ($x$-loss with SNR weighting) \\
Number of Steps & 300\,000 (rollout experiments) / 570\,000 (standard experiments) \\
EMA decay & 0.9999 \\
learning rate & 1e-4 \\
\bottomrule
\end{tabular}
\caption{Hyperparameters used for the Kinetics-600 experiments.}
\label{tab:hyperparams_k600}
\end{table}
\vspace{-2.5em}
\section{Runtime Complexity}
\label{sec:runtime_complexity}
Rolling diffusion models do not have an inherent runtime advantage over (batch) autoregressive diffusion models, as both have similar parameter sizes. 
For fair comparison, one can fix a number of \emph{evaluations-per-frame} for all models.
For example, we allow 32 model evaluations per frame.
An autoregressive model would sample a sequence frame by frame, each taking 32 evaluations.
A batch-autoregressive model samples, e.g., 8 frames jointly, allowing for 256 model evaluations for this subsequence.
In rolling diffusion using a window size of 8, the model gets evaluated 4 times before sliding the window. 
Upon sampling completion, every frame will be sampled using 32 evaluations.

Given this fixed number of model evaluations per frame, we show in this work that using a rolling sampling strategy can benefit time series generation.
However, we note that there are some slight inefficiencies during training.
Note that in rolling diffusion, some data is always revealed to the model due to the partial noising schedule. 
This means that if there is a high overlap between the frames, the model can directly copy the global structure from early frames to the later frames, achieving relatively good reconstruction loss.
We believe that because of this, the model gets a less strong signal for learning an ``image or video prior", as it can heavily rely on the conditioning signal.
Further experimentation to alleviate this issue is required, perhaps by adjusting the rolling noise schedule or fine-tuning a standard diffusion model using the rolling diffusion loss.

\section{Simulation Details}
\label{supp:simulation_details}
The parameters used to generate the Kolmogorov Flow data are shown in \Cref{table:hyperparameters_kolmogorov_sim}. It is important to note that to introduce uncertainty into an otherwise deterministic system, we vary the viscosity and density parameters, which must then be inferred from the data. This would also make a standard solver very difficult to use in such a setting. Additionally, due to the chaotic nature of the system, it is not deterministically predictable up to arbitrary precision. 
We use the `simple turbulence forcing' option, which combines a driving force with a damping term such that we simulate turbulence in two dimensions.

\begin{table}[H]
\centering
\caption{Parameters for Kolmogorov Flow Simulation using JaxCFD}
\label{table:hyperparameters_kolmogorov_sim}
\begin{tabular}{@{}ll@{}}
\toprule
Parameter             & Value                                  \\ \midrule
Size                  & 256                                    \\
Viscosity             & Uniform random in $[5.0 \times 10^{-4}, 5.0 \times 10^{-3}]$ \\
Density               & Uniform random in $[2^{-1}, 2^{1}]$    \\
Maximum Velocity      & 7.0                                    \\
CFL Safety Factor     & 0.5                                    \\
Max $\Delta t$            & 0.05                                 \\
Outer Steps           & 6000                                   \\
Grid                  & 256 $\times$ 256, domain $[0, 2\pi] \times [0, 2\pi]$ \\
Initial Velocity  & Filtered velocity field, 3 iterations \\
Forcing               & Simple turbulence, magnitude=2.5, linear coefficient=-0.1, wavenumber=4 \\
Total Simulations     & 200,000                                \\ \bottomrule
\end{tabular}
\end{table}

\section{Additional Results}
\label{supp:additional_results}
\subsection{Kolmogorov Flow}
We show in \Cref{tab:kolmogorov} the MSE and FSE errors at various time-steps.
Note that the MSE model is always optimal in terms of MSE loss, which is as expected.
However, in terms of matching the frequency distribution, as measured by FSD, standard diffusion, and in particular rolling diffusion are optimal. 
Averaging ensembles of the rolling diffusion model does not improve the FSD score, but does improve the MSE score.
\begin{table}[H]
    \centering
    \scalebox{0.8}{
    \begin{tabular}{@{}llllllll@{}}
      \toprule
                                           & \multicolumn{7}{c}{FSD/MSE @}                                                                                                                                                           \\
      Method                               & 1                             & 2                       & 4                      & 8                      & 12                      & 24                      & 48                      \\  \midrule
      MSE (2-1)                            & 304.7 / \textbf{14.64}        & 687.1 / \textbf{137.15} & 1007 / \textbf{407.0}  & 1649 / \textbf{407.0}  & 2230 / \textbf{407.0}   & 5453 / \textbf{407.0}   & 7504 / \textbf{407.0}   \\
      MSE (2-2)                            & 531.3 / 20.1                  & 7205 / 148.3            & 5596 / 407.0           & 7277 / 407.0           & 8996 / 406.1            & 1 $\cdot 10^4$ / 407.0  & 1 $\cdot 10^4$ / 407.0  \\
      MSE (2-4)                            & 304.7 / 21.7                  & 6684 / 148.9            & 6$\cdot10^4$ / 378.9   & 3 $\cdot 10^4$ / 407.0 & 2 $\cdot 10^4$ / 407.0  & $2 \cdot 10^4$ / 407.0  & 2 $\cdot 10^4$ / 407.0  \\  \midrule
      Standard Diffusion (2-1)             & 39.59 / 20.76                 & 57.73 / 192.6           & 142.0 / 710.0          & 297.7 / 794.6          & 399.4 / 773.1           & 442.5  / 758.3          & 763.6 / 732.5           \\
      Standard Diffusion (2-2)             & 59.15 / 47.88                 & 86.12 / 334.9           & 112.6 / 766.3          & 241.8 / 781.5          & 314.7 / 755.5           & 403.3  /  725.0         & 726.3 / 695.4           \\
      Standard Diffusion (2-4)             & 86.19 / 49.93                 & 141.6 / 353.6           & 246.6 / 753.2          & 397.8 / 758.0          & 555.6 / 726.9           & 1094.0 /  701.1         & 2401.0 / 666.3          \\
      Standard Diffusion (2-8)             & 87.0 / 54.0                   & 137.7 / 399.2           & 288.3 / 770.1          & 338.7 / 725.3          & 355.5 / 713.6           & 530.6 / 705.5           & 1159 / 748.3            \\ \midrule
      Rolling Diffusion (init noise) (2-2) & 63.21 / 45.62                 & 92.58 / 300.3  & 144.8 / 748.5 & 239.1 / 799.2 & 370.5 / 787.3 & 529.7 / 767.8   & 1568.7 / 735.3         \\
      Rolling Diffusion (init noise) (2-4) & 29.59 / 39.72                 & \textbf{47.44} / 287.8  & \textbf{43.39} / 738.4 & \textbf{61.93} / 769.0 & 214.32 / 735.7          & 648.53 / 699.1          & 1238.4 / 670.0         \\
      Rolling Diffusion (init noise) (2-8) & \textbf{27.68} / 41.22        & 52.41 / 316.9           & 53.47 / 768.0          & 98.29 / 777.2          & \textbf{187.03} / 748.8 & \textbf{344.89} / 737.6 & \textbf{417.59} / 719.4 \\ \midrule
      Rolling 10-Ensemble (2-8) & 481.6 / 36.4        & 8590 / 214.7           & 4 $\cdot 10^4$ / 429          & 5 $\cdot 10^4$ / 440          & 5 $\cdot 10^4$ / 429 & 5 $\cdot 10^4$ / 428 & 5 $\cdot 10^4$ / 429 \\
      \bottomrule
    \end{tabular}
    }
    \caption{Kolmogorov Flow  Results}
    \label{tab:kolmogorov}
  \end{table}
  
  \begin{figure}[H]
    \centering
    \includegraphics[width=0.49\linewidth]{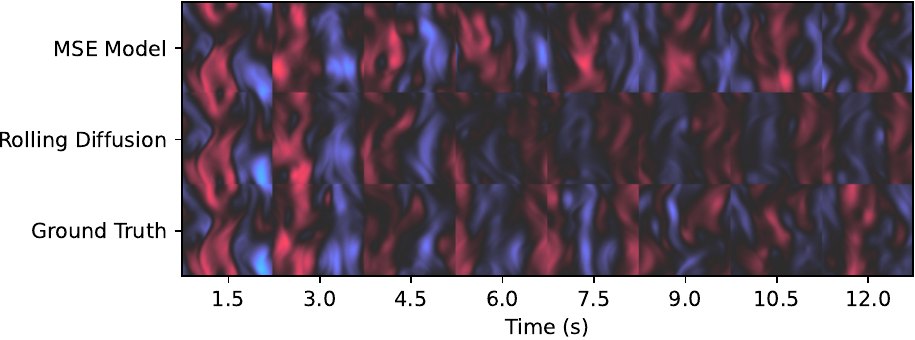} 
    \includegraphics[width=0.49\linewidth]{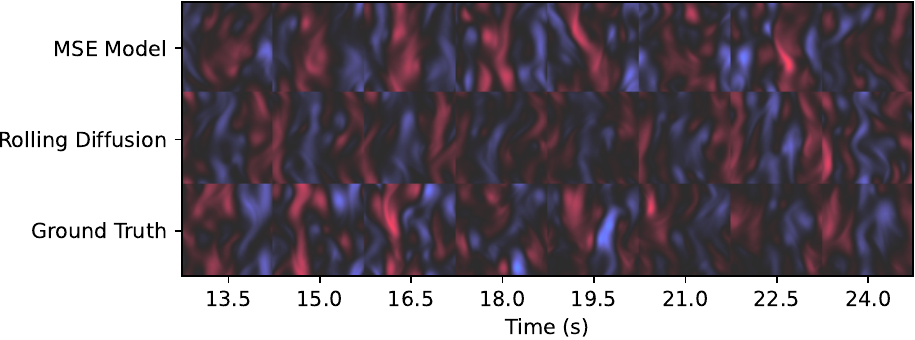} \\
    \includegraphics[width=0.49\linewidth]{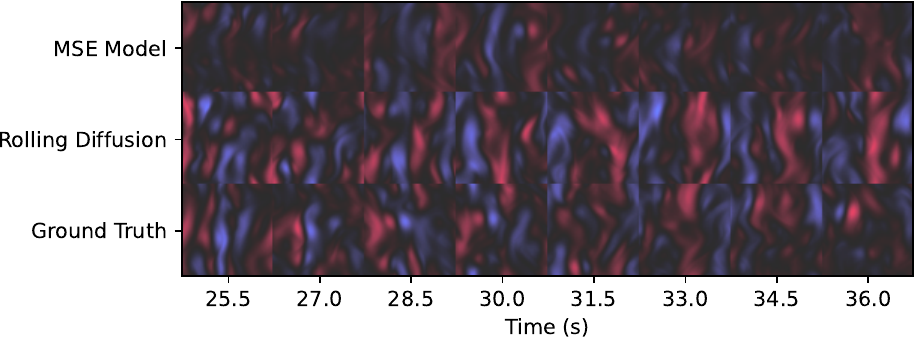} 
    \includegraphics[width=0.49\linewidth]{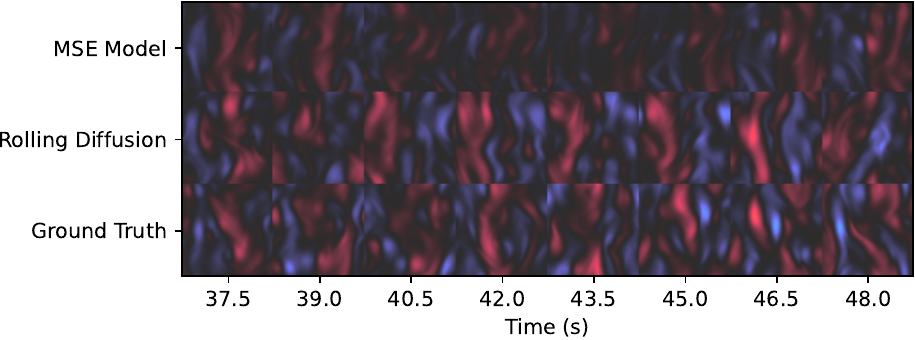} \\
    \includegraphics[width=0.49\linewidth]{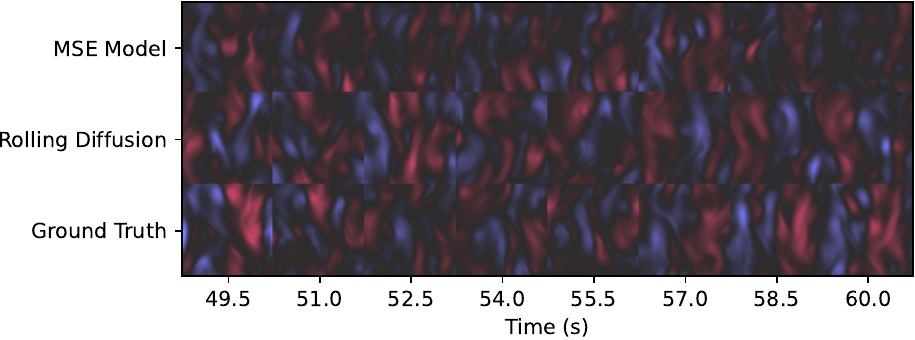} 
    \includegraphics[width=0.49\linewidth]{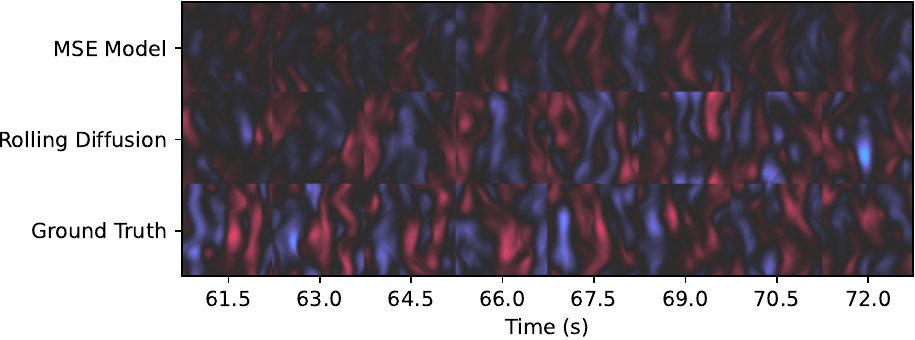} 
    \caption{Example rollout of a Kolmogorov flow sample. We depict the vertical velocity field. Note how the MSE model's intensity decreases as we move further from the conditioning frames.}
\end{figure}
  \subsection{Kinetics-600}
  In \Cref{tab:k600_1} we compare against TECO \citep{yan2023teco} on the Kinetics-600 dataset. 
  TECO uses 20 conditioning frames and generates 80 new frames at a resolution of 128 by 128 using 256 FVD samples.
\begin{table}[H]
  \centering
  \small

  \begin{tabular}{@{}lc@{}}
    \toprule
    Method                                         & FVD ($\downarrow$)     \\ \midrule
    TECO  & 799 \\
    Rolling Diffusion  (Ours)                            &    \textbf{685}              \\
    \bottomrule
  \end{tabular}
  \caption{In TECO \citep{yan2023teco}, samples are generated conditioning on 20 frames and generating 80 new frames at a resolution of 128 by 128 using 256 FVD samples. Different from other parts of the paper because of the low sample count, FVD is measured by \textit{matching} the conditioning for the reference samples.}
\end{table}

Furthermore, the following plot shows MSE deviations from ground-truth on Kinetics-600 data. 
\begin{figure}[H]
  \centering
  \includegraphics[width=0.5\textwidth]{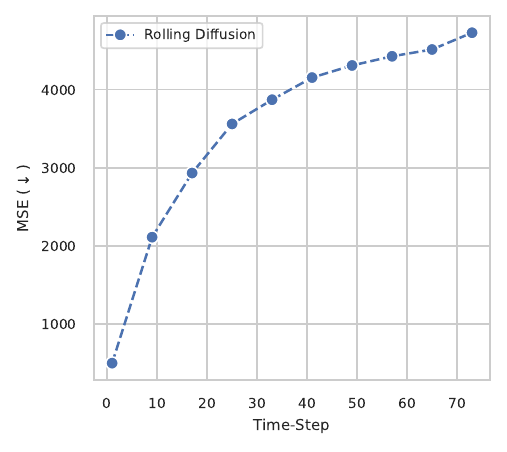}
  \caption{This figure shows MSE as a function of frame distance from the starting point on Kinetics-600 on the 20-12 setting with rollouts until frame 80 on resolution 128 $\times$ 128. As the generated frames are further from the initial 20 conditioning frames, the error between the original and the generated samples increases.}
  \label{fig:k600_exploding_error}
\end{figure}

\section{Rescaled Noise Schedule}
\label{sec:rescaled_schedule}
For our Kinetics-600 experiments, we used a different noise schedule which can sample from complete noise towards a ``rolling state'', i.e., at diffusion times $(\frac{1}{W}, \frac{2}{W}, \ldots \frac{W}{W})$.
From there, we can roll out generation using, e.g., the linear rolling sampling schedule $t_{w}^\lin$.
The reason is that we hypothesized that the noise schedule $t_{w}^\init$ uses a $\clip$ operation, which means that will be sampling in $\clean(s, t)$, which is redundant as outlined in the main paper and \Cref{sec:rolling_diffusion_objective}.

\begin{equation}
    t^{\mathrm{\init}, \mathrm{resc}}_w(t) := \frac{w}{W} + t \cdot (1 - \frac{w}{W})
\end{equation}
Where we clearly have at $t=\frac{1}{W}$ that the local times are $\left( \frac{1}{W}, \frac{2}{W}, \ldots, \frac{W}{W} \right)$, which is what we need.
Note that this schedule is not directly proportional to $t$.

\section{Hyperparameter Search for $\beta$}
\label{sec:beta_ablate}
\begin{table}[H]
  \small
  \centering

    \begin{tabular}{@{}llccc@{}}
      \toprule
     Task & Training regiment                                     & Cond-Gen & FVD            \\ \midrule
     \textit{stride=8}  & Rolling init rescaled {\small $\beta=0.1$} & (5-11)   & \textbf{39.8 } \\
     \textit{steps=24} & Rolling init rescaled  {\small $\beta=0.2$} & (5-11)   & 44.1               \\
      & Rolling init rescaled {\small $\beta=0.5$} & (5-11)   & 46.3               \\
      & Rolling init rescaled {\small $\beta=0.7$} & (5-11)   & 43.0           \\
      & Rolling init rescaled {\small $\beta=0.8$} & (5-11)   & 46.0           \\
      & Rolling init rescaled {\small $\beta=0.9$} & (5-11)   & 47.5           \\ 
      & Rolling init clip {\small $\beta=0.0$} & (5-11)   & 60.3               \\
      & Rolling init clip {\small $\beta=0.2$} & (5-11)   & 52.0           \\
      & Rolling init clip {\small $\beta=0.5$} & (5-11)   & 49.0           \\
      & Rolling init clip {\small $\beta=0.7$} & (5-11)   & 48.9           \\ 
      \midrule
   \textit{stride=1} & Rolling init rescaled {\small $\beta=0.7$} & (5-11) & 216  \\
   \textit{steps=64}  & Rolling init rescaled {\small $\beta=0.8$} & (5-11) & 227  \\
      & Rolling init rescaled {\small $\beta=0.9$} & (5-11)   & \textbf{211}              \\
      &   Standard Diffusion             & (15-1)   & 1460\hspace{.15cm} \\
     &   Standard Diffusion             & (5-1)   & 1369\hspace{.15cm} \\
     & Standard Diffusion             & (8-8)    & 157.1              \\
     & Standard Diffusion  & (5-8)   & 142.4 \\
      \bottomrule
    \end{tabular}
  \caption{Oversampling $t_w^\lin$ vs. $t_w^\init$ noise on Kinetics-600 with 8192 FVD samples. We allow 100 denoising steps per 11 frames.}
  \label{tab:sampling_training_overview}
\end{table}

\end{document}